\def\year{2022}\relax
\documentclass[letterpaper]{article} 
\usepackage{aaai22}  
\usepackage{times}  
\usepackage{helvet}  
\usepackage{courier}  
\usepackage[hyphens]{url}  
\usepackage{graphicx} 
\usepackage{natbib}  
\usepackage{caption} 
\DeclareCaptionStyle{ruled}{labelfont=normalfont,labelsep=colon,strut=off} 
\usepackage{algorithm}
\usepackage{algorithmic}

\usepackage{amsmath}
\usepackage{amsfonts,amssymb}
\usepackage{mathrsfs}
\usepackage{booktabs}
\usepackage{multirow}
\usepackage{makecell}

\usepackage{pgfplots}

\usepackage{newfloat}
\usepackage{listings}
\floatstyle{ruled}
\newfloat{listing}{tb}{lst}{}
\floatname{listing}{Listing}
\title{Learn from Yesterday: A Semi-Supervised Continual Learning Method for Supervision-Limited Text-to-SQL Task Streams}
\author{
    Yongrui Chen\textsuperscript{\rm 1},
	Xinnan Guo\textsuperscript{\rm 1},
	Tongtong Wu\textsuperscript{\rm 1,2},
	Guilin Qi\textsuperscript{\rm 1},
	Yang Li\textsuperscript{\rm 3},
	Yang Dong\textsuperscript{\rm 4}
	\\
}
\affiliations{
    \textsuperscript{\rm 1}School of Computer Science and Engineering, Southeast University\\
    \textsuperscript{\rm 2}Department of Data Science \& AI, Monash University\\
    \textsuperscript{\rm 3}Alibaba Group \quad
    \textsuperscript{\rm 4}Ant Group\\

    yrchen@seu.edu.cn, guoxinnan0727@163.com,
    wutong8023@seu.edu.cn, tongtong.wu@monash.edu\\
    gqi@seu.edu.cn, ly200170@alibaba-inc.com, dongyang1231@gmail.com
}

\usepackage{bibentry}

\begin{document}

\maketitle


\begin{abstract}
Conventional text-to-SQL studies are limited to a single task with a fixed-size training and test set. When confronted with a stream of tasks common in real-world applications, existing methods struggle with the problems of insufficient supervised data and high retraining costs.
The former tends to cause overfitting on unseen databases for the new task, while the latter makes a full review of instances from past tasks impractical for the model, resulting in forgetting of learned SQL structures and database schemas.
To address the problems, this paper proposes integrating \textit{semi-supervised learning} (SSL) and \textit{continual learning} (CL) in a stream of text-to-SQL tasks and offers two promising solutions in turn. The first solution \textsc{Vanilla} is to perform self-training, augmenting the supervised training data with predicted pseudo-labeled instances of the current task, while replacing the full volume retraining with episodic memory replay to balance the training efficiency with the performance of previous tasks.
The improved solution \textsc{SFNet} takes advantage of the intrinsic connection between CL and SSL. It uses in-memory past information to help current SSL, while adding high-quality pseudo instances in memory to improve future replay. 
The experiments on two datasets shows that \textsc{SFNet} outperforms the widely-used SSL-only and CL-only baselines on multiple metrics.
\end{abstract}

\section{Introduction}
Relational databases (RDs) store a vast amount of today's information and provide the foundation for applications such as customer relationship management~\cite{anshari2019customer}, financial markets~\cite{lewis2017blockchain}, and medical records~\cite{wang2020text}. Text-to-SQL technology trains a parser to translate natural language problems into machine-readable SQL programs, providing an ideal way for non-technical users to easily interact with their data stored in RDs. Current research on text-to-SQL has covered single-table~\cite{DBLP:journals/corr/abs-1709-00103}, multi-table~\cite{DBLP:conf/emnlp/YuZYYWLMLYRZR18}, and conversation~\cite{DBLP:conf/emnlp/YuZELXPLTSLJYSC19} scenarios, with a common assumption that the size of training and testing data does not change over time. Unfortunately, in real-world applications, new databases are always emerging to adapt to changing circumstances (e.g., new diseases, adjusted financial policies), thus continuously generating new tasks for the parser. Although machine learning-based text-to-SQL methods have achieved state-of-the-art (SOTA) performance, they suffer from the following two challenges in the face of rapidly growing tasks.

\begin{figure}
	\includegraphics[width=0.48\textwidth]{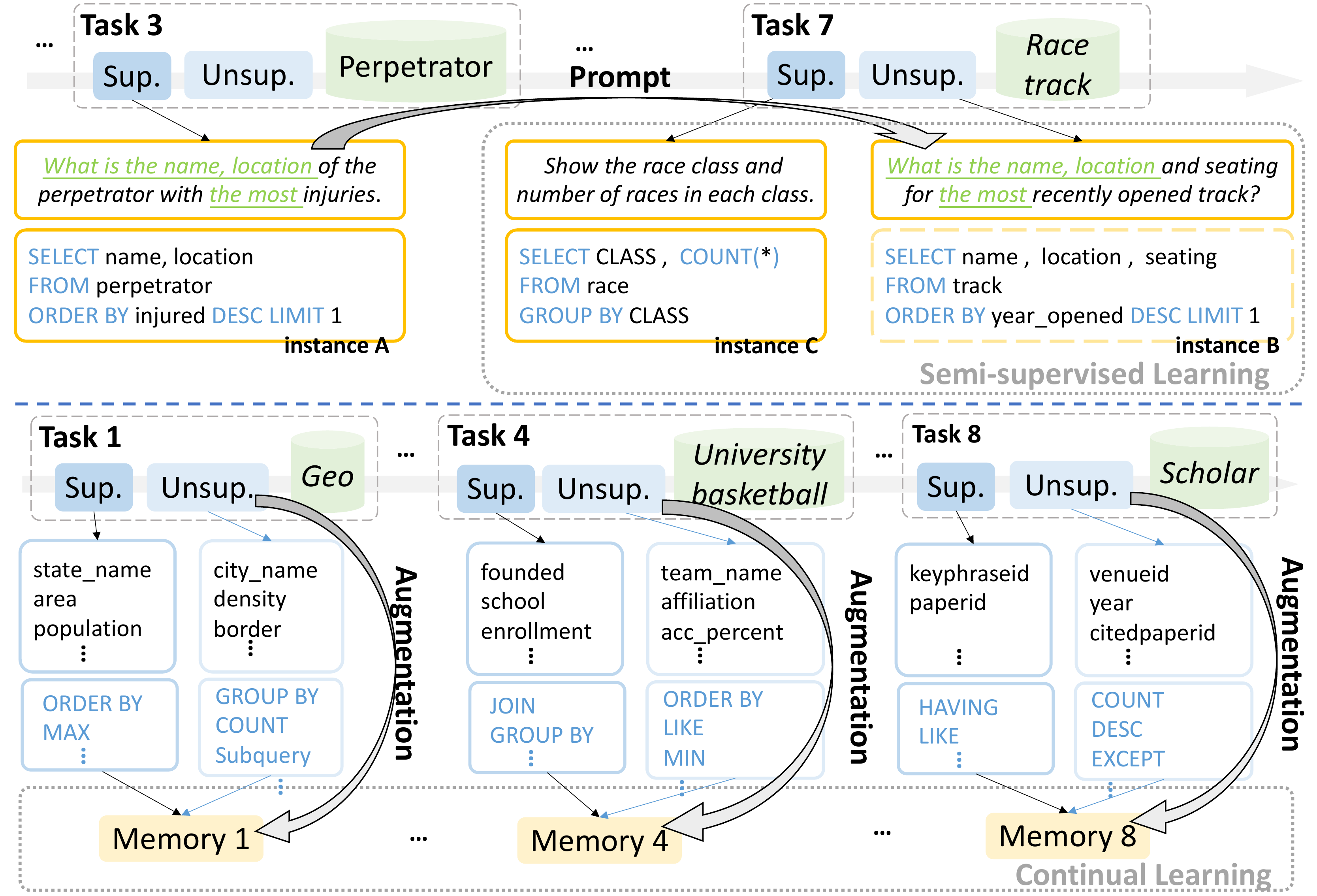}
	\caption{Mutual promotion of SSL and CL. \texttt{Sup} denotes labeled data, \texttt{Unsup} denotes unlabeled data, green cylindrical denotes the database against the task.} \label{fig_example}
\end{figure}

1) Limited supervised data. For a new text-to-SQL task against unseen databases, it is often impractical to annotate sufficient SQL labels for training in a short period of time, resulting in parsers that are prone to overfitting.
2) Costly full volume retraining. Considering a new task, an intuitive idea is to train the model from scratch on all seen tasks. Unfortunately, the computational cost of such retraining is unaffordable due to the increasing size of pre-trained models, even with limited training data~\cite{wu2022pretrained}. Assuming that there are $K$ tasks and the average training time of one task is $T_{\text{avg}}$, the total time of full volume retraining on these $K$ tasks is $T=K(K+1)T_{\text{avg}}/2$. When $K$ is large, the squared relationship makes $T$ intolerable.


Existing works provide ideas to address these two challenges separately. First, \textit{semi-supervised learning} (SSL)~\cite{DBLP:conf/ijcai/Guo0QWX22} can compensate for the lack of supervised data by mining the potential value of NLQs without SQL labels. Second, \textit{continuous learning} (CL)~\cite{DBLP:conf/emnlp/LiQH21} provides an alternative cost-effective training paradigm without using all instances of previous tasks. However, both of them focuses on one of the challenges but ignores the impact of the other. For text-to-SQL task streams, it remains a pressing issue to address both challenges simultaneously.

In this paper, we propose to integrate SSL and CL to solve the supervision-limited text-to-SQL task stream. We first give a \textsc{Vanilla} solution in which the parser applies self-training to predict pseudo-labeled instances to improve generalization to the current task (SSL) while replaying small portions of past instances stored in memory to alleviate forgetting of previous tasks (CL). Despite its simplicity, \textsc{Vanilla} has experimentally proven to be sufficient to outperform most SSL-only and CL-only methods in text-to-SQL task streams. Thereafter, we further hypothesize that SSL and CL can boost each other. On the one hand, some instances in previous tasks can supply valuable information to SSL for predicting pseudo-labels of unlabeled instances. As shown in Figure~\ref{fig_example}(a), although instances $A$ and $B$ are associated with different databases, \texttt{Perpetrator} and \texttt{Race\_track}, respectively, the target SQL of $A$ is similar to that of $B$. The parser might learn from $A$ on how to predict the pseudo-label of $B$. On the other hand, high-quality pseudo-labeled instances can also enrich the memory of past tasks. In the three tasks shown in Figure~\ref{fig_example}(b), the pseudo instances can supplement the supervised-only memories in terms of both RD schemas and SQL keywords.
Motivated by this, we propose the \textit{soft fusion network} (\textsc{SFNet}), which applies a \textit{teacher-student framework} to separately cope with SSL and CL processes. Specifically, \textsc{Teacher} is committed to the optimum on each single task via self-training, while \textsc{Student} learns the pseudo labels predicted by \textsc{Teacher} on all seen tasks via replaying to achieve the optimum on the entire task stream.
To utilize the mutual promotion of CL and SSL, \textsc{SFNet} performs \textit{dual sampling}: when training \textsc{Teacher}, past instances relevant to the current task are used to prompt the SSL process; when training \textsc{Student}, both labeled and unlabeled instances of previous tasks are sampled to guarantee complete memory for replay.
Comprehensive experiments on two text-to-SQL benchmarks show that our SFNet further improves on the \textsc{Vanilla} solution, achieving SOTA results on multiple metrics. In summary, the contributions of this paper include:
\begin{itemize}
    \item We propose a \textsc{Vanilla} solution that combine SSL and CL to solve the problem of supervision-limited text-to-SQL task stream. To the best of our knowledge, this is the first time that the two technologies have been integrated in text-to-SQL.
    
    \item Based on the \textsc{Vanilla} solution, we propose an improved solution \textsc{SFNet} that applies a teacher-student framework to isolate the different optimization goals of SSL and CL, and employs a dual sampling strategy to exploit their mutual facilitation.
    
    \item We covert two mainstream text-to-SQL datasets into the form of task streams and conduct comprehensive experiments. Our methods outperform the existing CL-only and SSL-only competitors and achieves SOTA performance on multiple metrics.
\end{itemize}

\section{Preliminaries}

\subsection{A Base Text-to-SQL Parser}
Given an NLQ $q$ and a schema $\mathcal{S}=(\mathcal{C}, \mathcal{T})$ for an RD, conventional text-to-SQL aims to generate a SQL program $y$ by 
\begin{equation}
    y = \mathcal{F}_{\theta}(q,\mathcal{S}),
\end{equation}
where $\mathcal{S}$ consists of columns $\mathcal{C}=\{c_1,...,c_{|\mathcal{C}|}\}$, tables $\mathcal{T} = \{t_1,...t_{|\mathcal{T}|}\}$, and $\mathcal{F}_{\theta}$ denotes a parser with parameters $\theta$. 
Most SOTA text-to-SQL parsers for the single task~\cite{DBLP:conf/acl/GuoZGXLLZ19,DBLP:conf/acl/WangSLPR20,DBLP:conf/acl/CaoC0ZZ020} represent the desired $y$ as an \textit{abstract syntax tree} (AST) $\mathcal{Z}_y$ via a context-free grammar and adopt a \textit{sequence-to-sequence} architecture~\cite{luong2015effective} to synthesize $\mathcal{Z}_y$. To make a fair comparison between different learning algorithms, we built a robust parser $\mathcal{F}_{\theta}$ along their lines.

\subsubsection{Encoder} 
An strong table pre-trained model \textsc{Grappa}~\cite{DBLP:conf/iclr/0009WLWTYRSX21} is employed to encode $q$ and $\mathcal{S}$ into a sequence of contextual word representations, $\mathbf{q}$ and $\mathbf{S}$. It was pre-trained on an 866.5k table-text corpus, being injected with structural properties common to the semantic parsing of tables.

\subsubsection{Decoder} 
A \textit{long short-term memory} (LSTM) network is used as a decoder to synthesize $\mathcal{Z}_y$ by generating a sequence of actions $Z(y) = \{z_1, z_2...z_{|Z(y)|}\}$. These actions determine the SQL skeleton and the used schemas according to a top-down grammar, \textsc{SemQL}~\cite{DBLP:conf/acl/GuoZGXLLZ19}. In particular, the probability of $\mathcal{Z}_y$ is estimated by
\begin{equation}
    P(\mathcal{Z}_y|\mathbf{q}, \mathbf{S}, \theta) = {\prod_{j=1}^{|Z(y)|} P(z_{j}|\mathbf{q}, \mathbf{S}, z_{< j}, \theta)},
\end{equation}
where $z_{j}$ is the $j$-th action. $P(z_{j}|q, \mathcal{S}, z_{< j})$ denotes the predicted probability of $z_{j}$, determined by the normalized inner product of the hidden state ${\mathbf{h}}_j$ of the LSTM and the embeddings of candidate actions.

\subsection{Problem Formulation}
In our scenarios, $\mathcal{F}_{\theta}$ is trained continually by a sequence of $K$ distinct tasks $\{\mathcal{D}^{1},\mathcal{D}^{2},...,\mathcal{D}^{K}\}$. Each task $\mathcal{D}^{i}$ consists of a labeled training set $\mathcal{A}^{i}=\{a_1^i...,a_{|\mathcal{A}^{i}|}^i\}$, a validation set $\mathcal{D}^{i}_{\text{valid}}$, and a test set $\mathcal{D}^{i}_{\text{test}}$, where each training instance $a_k^i=(q_k^i, \mathcal{S}_k^i, \tilde{y}_k^i)$ has a gold SQL program $\tilde{y}_k^i$ as the label.
Considering the limited supervision, we let $|\mathcal{A}^{i}|$ far smaller than the size of the general text-to-SQL datasets. In addition, we also assume that there is an unlabeled set $\mathcal{U}^{i}=\{u_1^i...,u_{|\mathcal{U}^{i}|}^i\}$ for $\mathcal{D}^{i}$, where $u_k^i=(q_k^i, \mathcal{S}_k^i)$ does not have gold SQL labels. Different tasks depend on different RDs, i.e., for $\forall \mathcal{D}^i, \forall \mathcal{D}^j,$ if $i \neq j$, then $S(\mathcal{D}^i) \cap S(\mathcal{D}^j) = \emptyset$, where $S(\mathcal{D}^i)$ denotes the set of corresponding RDs. 
Our final goal is to make $\mathcal{F}_{\theta}$ achieve a good SQL generation accuracy on each $\mathcal{D}^{i}_{\text{test}}$ after the training of all $K$ tasks.

\section{A Vanilla Solution}
\begin{figure}
	\includegraphics[width=0.49\textwidth]{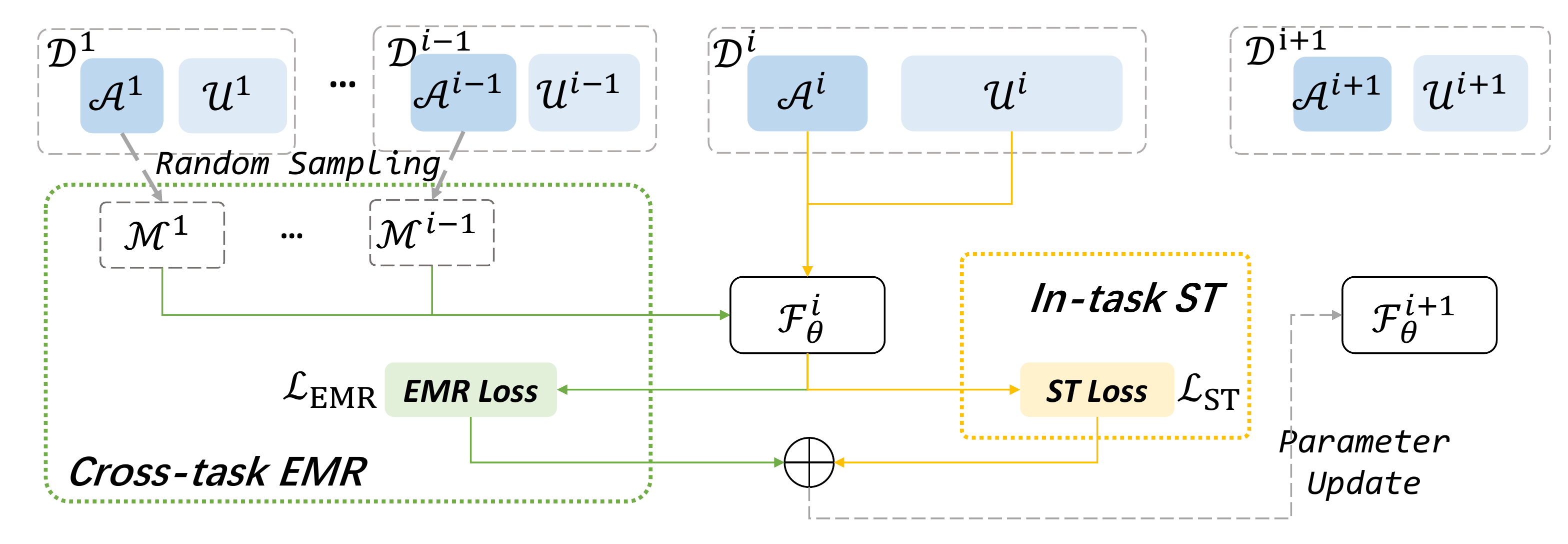}
	\caption{Architecture of our proposed \textsc{Vanilla} solution.} \label{fig:vanilla}
\end{figure}

We start with a \textsc{Vanilla} solution to overcome the challenges of training $\mathcal{F}_{\theta}$ on $\{\mathcal{D}^{1},\mathcal{D}^{2},...,\mathcal{D}^{K}\}$, consisting of \textit{in-task self-training} and \textit{cross-task episodic memory replay}. Figure \ref{fig:vanilla} shows its entire architecture.

\subsection{In-task Self-training}
As a classical SSL method, \textit{self-training} (ST) uses the model's predictions to product pseudo-labeled data to augment the limited supervision (the first challenge). 
Inspired by this, during the learning of task $\mathcal{D}^i$, we adopt a two-stage strategy to improve the generalization capability of $\mathcal{F}_{\theta}$. 

\subsubsection{Warm Start} In the first stage, $\mathcal{F}_{\theta}^i$ is trained on labeled set $\mathcal{A}^i$ for multiple epochs to fully understand $\mathcal{D}^i$, thus guaranteeing the quality of the first batch of pseudo labels predicted by $\mathcal{F}_{\theta}^i$. Specifically, for each labeled instance $a_k^i = (q_k^i, \mathcal{S}_k^i, \tilde{y}_k^i) \in \mathcal{A}^i$, the loss is calculated by a log-likelihood,
\begin{equation}
    \mathcal{L}(a_k^i;\theta^i) =
    -{\sum_{j=1}^{|Z|} \log P(\tilde{z}_{j}|\mathbf{q}_k^i, \mathbf{S}_k^i, z_{< j}^i, \theta^i)}.
\end{equation}

\subsubsection{Self-updating} In the second stage, $\mathcal{F}_{\theta}^i$ is trained on $\mathcal{A}^i \cup \mathcal{U}^i$. At each epoch, it first predicts a SQL program $\hat{y}_k$ for each unlabeled instance $u^i_k \in \mathcal{U}^i$ to a pseudo-labeled instance $p^i_k=(q_k^i, \mathcal{S}_k^i, \hat{y}_k^i)$. Random $k$ instances $p^i_k$ are selected to compose the pseudo-label set $\mathcal{P}^i$. Subsequently, $\mathcal{F}_{\theta}$ is updated by optimizing
\begin{equation}
    \mathcal{L}_{\text{ST}}(\mathcal{A}^i \cup \mathcal{P}^i;\theta^i) = \sum_{a^i_k \in \mathcal{A}^i} \mathcal{L}(a^i_k; \theta^i) + \sum_{p^i_k \in \mathcal{P}^i} \mu_k \mathcal{L}(p^i_k; \theta^i),
\end{equation}
where $\mu_k=P(\mathcal{Z}_y|\mathbf{q}, \mathbf{S}, \theta)$ is the confidence score to evaluate the contribution of each $p^i_k$ to the loss.

\subsection{Cross-task Episodic Memory Replay}
Considering the second challenge,  $\mathcal{F}_{\theta}$ cannot be retrained with the full set of previous instances when a new task is encountered. This also brings up a new problem of catastrophic forgetting (CF), i.e., $\mathcal{F}_{\theta}$ forgets the past tasks after learning the new ones. Fortunately, \textit{episodic memory replay} (EMR) can balance the efficiency and CF by allowing the model to review a portion of the experienced instances. In addition, compared with other gradient-based CL methods~\cite{DBLP:journals/corr/KirkpatrickPRVD16,DBLP:conf/nips/Lopez-PazR17}, EMR is well suited for  the text-to-SQL task with complex labels $Z(y)$ because of its simple process. Therefore, we add the following process along with ST.
\subsubsection{Memory Construction}
As a preparation, we construct a fixed-size memory $\mathcal{M}^i=\{m_1^i, m_2^i, ..., m_{|\mathcal{M}^i|}^i\}$ associated with each $\mathcal{D}^i$ to store a small number of replay instances, where $m_k^i = (q_k^i, \mathcal{S}_k^i, \tilde{y}_k^i)$ is sampled by from $\mathcal{A}^{i}$. This setup is practical for text-to-SQL parsers because they have difficulty recalling information from the past except for reusing the instances stored in memory. 
\subsubsection{Replay Loss}
Concretely, whenever $\mathcal{F}_{\theta}^i$ performs self-training, $\mathcal{L}_{\text{EMR}}$, the loss of all replayed instances stored in $\mathcal{M}^1$ to $\mathcal{M}^{i-1}$, is added to $\mathcal{L}_{\text{ST}}$,
\begin{equation}
    \mathcal{L}_{\text{EMR}}(\mathcal{M}^j;\theta^i) = \sum_{j=1}^{i-1} \sum_{m_k^j \in \mathcal{M}^j} {\mathcal{L}(m_k^j; \theta^i)}.
\end{equation}
Following the common practice, at the end of the training of $\mathcal{D}^{i}$, we randomly select a $M$ labeled instances from $\mathcal{A}^i$ and store them in $\mathcal{M}^{i}$ for replay in future tasks.


\section{Soft Fusion Network}
\begin{figure}
	\includegraphics[width=0.49\textwidth]{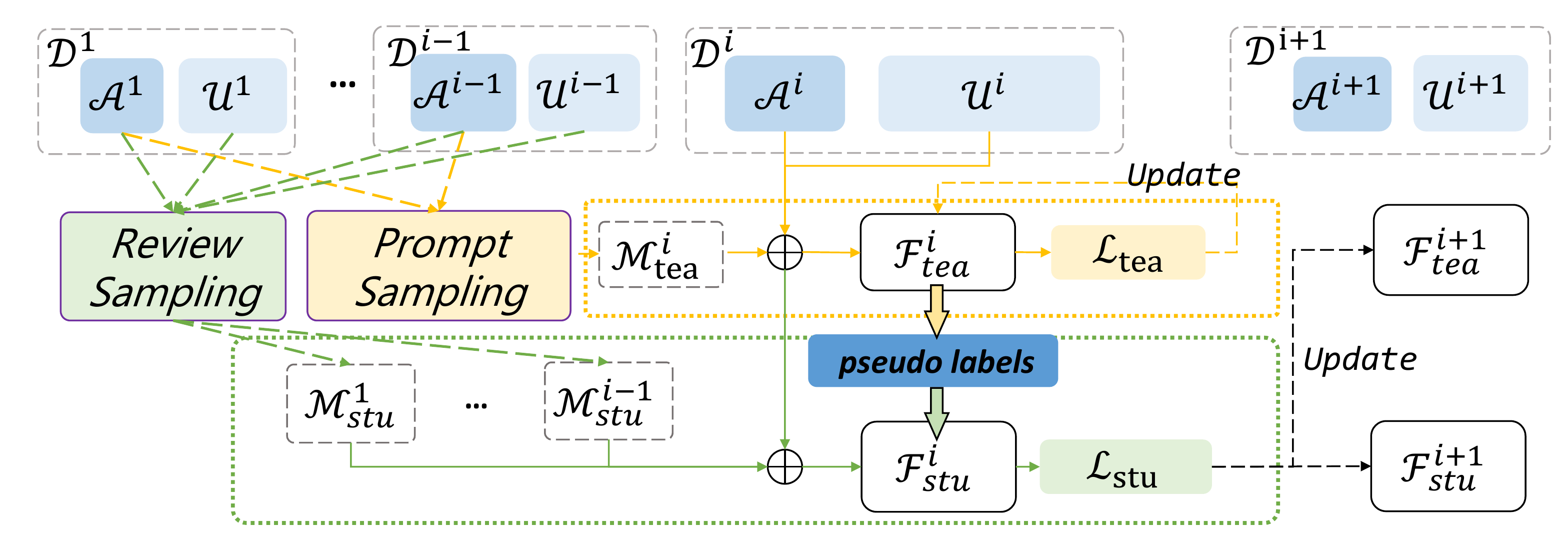}
	\caption{Architecture of our proposed \textsc{SFNet}.} \label{fig_sfnet}
\end{figure}

Figure \ref{fig_sfnet} shows the architecture of our proposed \textsc{SFNet}, which is an improved version of \textsc{Vanilla}. Since SSL is dedicated to the optimization of a single (current) task, while CL is more concerned with the overall performance of all tasks, \textsc{SFNet} applies a \textit{Teacher-Student Framework} (TS) to perform them separately. It composes of two base parsers, \textsc{Teacher} $\mathcal{F}_{\text{tea}}$ for SSL and \textsc{Student} $\mathcal{F}_{\text{stu}}$ for CL. During the task $D^i$, both $\mathcal{F}_{\text{tea}}^i$ and $\mathcal{F}_{\text{stu}}^i$ are initialized from $\mathcal{F}_{\text{stu}}^{i-1}$ but with separate parameter updates during each task. To drive the mutual promotion of SSL and CL (Figure \ref{fig_example}), \textsc{SFNet} uses \textit{dual sampling} that contains two different strategies to augment the training data of $\mathcal{F}^i_{\text{tea}}$ and $\mathcal{F}^i_{\text{stu}}$, respectively.

\subsection{Teacher-Student Framework}
\subsubsection{\textsc{Teacher} Parser}
The sole goal of $\mathcal{F}^i_{\text{tea}}$ is to offer correct pseudo labels of the current task $\mathcal{D}^i$ via the in-task ST. Therefore, it only focuses on $\mathcal{D}^i$ and is allowed to forget past tasks that is not associated with $\mathcal{D}^i$. In contrast, previous instances relevant to $\mathcal{D}^i$ can be highlighted in order to deepen $\mathcal{F}^i_{\text{tea}}$'s understanding of $\mathcal{D}^i$. To achieve this goal, during the ST, we refer to EMR to replay the appropriate instances having potential prompts. In particular, when observing each task $\mathcal{D}^i$, $N$ labeled instances relevant to $\mathcal{D}^i$ are drawn from $\bigcup_{k=1}^{i-1} \mathcal{A}^k \cup \mathcal{P}^k$ to compose the memory $\mathcal{M}_{\text{tea}}^i=\{m_1^i...,m_N^i\}$. We call this step \textit{prompt sampling} and detail it in the subsequent section. Thereafter, $\mathcal{F}^i_{\text{tea}}$ follows \textsc{Vanilla} to perform in-task ST except that it replaces $\mathcal{L}_{\text{ST}}$ in self-updating with the following loss,
\begin{equation}
    \mathcal{L}_{\text{tea}}(\theta^i_{\text{tea}}) = \mathcal{L}_{ST}(\mathcal{A}^i \cup \mathcal{P}^i;\theta^i_{\text{tea}}) + \sum_{m^i_k \in \mathcal{M}^i_{\text{tea}}} \mathcal{L}(m^i_k; \theta_{\text{tea}}^i).
\end{equation}
When the training converges, $\mathcal{F}^i_{\text{tea}}$ is considered to be approaching the optimum of $\mathcal{D}^i$. Note that the cost here is that $\mathcal{F}^i_{\text{tea}}$ may forget some key information of $\mathcal{D}^1...,\mathcal{D}^{i-1}$.

\subsubsection{\textsc{Student} Parser}
Assuming that trained $\mathcal{F}^i_{\text{tea}}$ is an expert proficient in $\mathcal{D}^i$, then the task stream could theoretically provide $K$ experts for $\mathcal{D}^1...,\mathcal{D}^{K}$. Intuitively, if a parser can inherit the capabilities of all these experts, then it will be overall optimal of the entire task stream. Our $\mathcal{F}^i_{\text{stu}}$ aims to be such a parser that for each task $\mathcal{D}^{j}$ ($1 \le j \le i$) learns from the pseudo-labeled instances $\mathcal{P}^j$ generated by the trained $\mathcal{F}^j_{\text{tea}}$ in addition to the original $\mathcal{A}^j$. Considering the training efficiency, our practice is that using cross-task EMR during the learning of each $\mathcal{D}^i$.
Concretely, the loss of $\mathcal{F}^i_{\text{stu}}$ contains 1) the task loss on $\mathcal{D}^i$ and 2) the replay loss on $\mathcal{D}^1...,\mathcal{D}^{i-1}$,
\begin{equation}
\begin{aligned}
    \mathcal{L}_{\text{stu}}(\theta^i_{\text{stu}}) = \sum_{x^i_k \in {\mathcal{A}^i \cup \mathcal{P}^i}}  \mathcal{L}(x^i_k; \theta_{\text{stu}}^i)
    + \mathcal{L}_{\text{EMR}}(\mathcal{M}^j_{\text{stu}};\theta_{\text{stu}}^i),
\end{aligned}
\end{equation}
where the replayed instances $\mathcal{M}^j_{\text{stu}} = \{m_1^j...,m_M^j\}$ ($1 \le j \le i$) are sampled from $\mathcal{A}^j \cup \mathcal{P}^j$ via \textit{review sampling} (detailed in the next section). The sampled instances are diversified in both the SQL skeleton and the RD schema so that $\mathcal{F}^i_{\text{stu}}$ can recall complete information about past tasks. 

\subsection{Dual Sampling}

\subsubsection{Prompt Sampling}
x

Algorithm \ref{alg:teacher_sampling} details the process of prompt sampling (PS). First, $\textsc{TopK}(\mathcal{R}, \Omega, \tilde{N})$ denotes that selecting $\tilde{N}$ instances $x_j$ from $\mathcal{R} = \bigcup_{k=1}^{i-1} \mathcal{A}^k \cup \mathcal{P}^k$ to form a temporary memory $\tilde{\mathcal{M}}_{\text{tea}}$ according to the relevance score $\Omega = \{\omega(x_1),...,\omega(x_{|\mathcal{R}|})\}$, 
\begin{equation}
\omega(x_j) = \max_{x_l \in \mathcal{U}^i} d_{\text{sch}}(x_l, x_j),
\end{equation}
where $d_{\text{sch}}(x_l, x_j) = \sqrt{\sum_{\psi \in \Psi}(\mathbf{v}_{\psi}^l - \mathbf{v}_{\psi}^j)^2}$ denotes the distance of $x_l$ and $x_j$ on the schemas $\mathcal{S}^k$ and $\mathcal{S}^j$. 
Here $\Psi$ is the vocabulary of schema tokens and $\mathbf{v} \in \mathbb{R}^{|\Psi|}$ is the hash vector of $\mathcal{S}$. Specifically, if the token $\psi$ exists in $y$, then $\mathbf{v}_{\psi}$ is 1, otherwise it is 0. Thereafter, $\tilde{\mathcal{M}}_{\text{tea}}$ is partitioned into $n$ clusters with structure distance $d_{\text{stru}}$ to meet the diversity of SQL skeletons. $d_{\text{stru}}$ is formally similar to $d_{\text{sch}}$, but replacing $\Psi$ with the SQL keywords vocabulary $\Phi$ (including \texttt{GROUP BY}, \texttt{LIMIT}, etc.). The reason for using hashing techniques for $d_{\text{stru}}$ and $d_{\text{sch}}$ is to simplify the process. More complex distance metrics are left for future work.
Finally, the central instance of each cluster is selected to compose $\mathcal{M}_{\text{tea}}$. Note that although our PS requires re-traversing all past instances for each new task, its time is only a fraction of the total time of \textsc{SFNet} in our experiments.

\subsubsection{Review Sampling} 
Review sampling (RS) is an advanced version of the random sampling in cross-task EMR of \textsc{Vanilla}. The most significant difference between them is that RS samples pseudo-labeled instances $\mathcal{P}^i$ in addition to $\mathcal{A}^i$, where the labels of each $p \in \mathcal{P}^i$ is predicted by $\mathcal{F}^i_{\text{tea}}$ via the in-task ST. In this way, the resulting memory is augmented, allowing $\mathcal{F}^i_{\text{stu}}$ to recall the task $\mathcal{D}^i$ more fully.
Naturally, we would like sampled instances to be representative of $\mathcal{D}^i$ in terms of both SQL skeletons and RD schemas. Thus, we define a combined distance $d(x_1, x_2) = d_{\text{stru}} * d_{\text{sch}}$ and use it to partition  $\mathcal{A}^i \cup \mathcal{P}^i$ into $M$ clusters. Here we use the product in order to balance the weights of $d_{\text{stru}}$ and $d_{\text{sch}}$. Consistent with PS, the result $\mathcal{M}^j_{\text{stu}}$ consists of all central instances of the cluster, which can be considered as the representative of $\mathcal{D}^i$.

\begin{algorithm}[t]
\setlength\belowdisplayskip{1pt}
	\caption{Prompt Sampling}
	\begin{algorithmic}[1]
		\label{alg:teacher_sampling}
		\REQUIRE Past labeled instances set  $\mathcal{R} = \bigcup_{k=1}^{i-1} \mathcal{A}^k \cup \mathcal{P}^k$, current unlabeled set $\mathcal{U}^i$, score $\mathbf{\Omega}=\{\omega(x_1),...,\omega(x_{|\mathcal{R}|})\}$
        \STATE $\tilde{\mathcal{M}}_{\text{tea}} \leftarrow \textsc{TopK}(\mathcal{R}, \Omega, \tilde{N})$, $\mathcal{M}_{\text{tea}} \leftarrow \emptyset$
		\STATE Partition $\tilde{\mathcal{M}}_{\text{tea}}$ into $N$ clusters, denoted as $C$, with the SQL structure distance metric $d_{\text{stru}}$
		\FOR{$c_j \in C$}
		\STATE $\mathcal{M}_{\text{tea}} \leftarrow \mathcal{M}_{\text{tea}} \cup \{(q^*, \mathcal{S}^*, y^*)\}$, where $(q^*, \mathcal{S}^*, y^*)$ is closest to the center of cluster $c_j$
		\ENDFOR
		\RETURN $\mathcal{M}_{\text{tea}}$
	\end{algorithmic}
\end{algorithm}
\begin{figure}
\centering
    \begin{tikzpicture}[scale=0.5]
        \begin{axis}[
        ybar,
        bar width=1.5pt,
        enlargelimits=0.15,
        ylabel=Seconds,
    	xtick=data,
        ymajorgrids=true, 
        grid style=dashed]
    	
        \addplot[draw=blue, fill=blue!60!white]
    	coordinates {(1, 788) (2, 750) (3, 734) (4, 498) (5, 476) (6, 420) (7, 574) (8, 326) (9, 432) (10, 330)};
    	
    	\addplot[draw=cyan, fill=cyan!30!white] 
    	coordinates {(1, 423) (2, 877) (3, 870) (4, 748) (5, 736) (6, 708) (7, 790) (8, 636) (9, 717) (10, 684)};
    
        \addplot[draw=yellow, fill=yellow!60!white]
    	coordinates {(1, 574) (2, 1014) (3, 1008) (4, 809) (5, 802) (6, 767) (7, 877) (8, 662) (9, 771) (10, 695)};
    	
    	\addplot[draw=brown, fill=brown!30!white]
    	coordinates {(1, 245) (2, 664) (3, 694) (4, 610) (5, 578) (6, 556) (7, 589) (8, 502) (9, 568) (10, 529)};
    
        \end{axis}
    \end{tikzpicture}
    \begin{tikzpicture}[scale=0.5]
        \begin{axis}[
        ybar,
        bar width=1.5pt,
        enlargelimits=0.15,
        legend style={ at={(0.95,0.95)}},
        ylabel=Seconds,
    	xtick=data,
        ymajorgrids=true, 
        grid style=dashed]
    	
        \addplot[draw=blue, fill=blue!60!white]
    	coordinates {(1, 967) (2, 520) (3, 332) (4, 159) (5, 237) (6, 181) (7, 222) (8, 180) (9, 242) (10, 304)};
    	\addlegendentry{$|\mathcal{A}^{i}|$}
    	
    	\addplot[draw=cyan, fill=cyan!30!white] 
    	coordinates {(1, 1224) (2, 586) (3, 345) (4, 185) (5, 339) (6, 265) (7, 281) (8, 284) (9, 338) (10, 357)};
    	\addlegendentry{$|\mathcal{U}^{i}|$}
    	
    	\addplot[draw=yellow, fill=yellow!60!white]
    	coordinates {(1, 347) (2, 205) (3, 152) (4, 98) (5, 145) (6, 106) (7, 91) (8, 87) (9, 135) (10, 80)};
    	\addlegendentry{$|\mathcal{D}^{i}_{\text{test}}|$}
    	
    	\addplot[draw=brown, fill=brown!30!white]
    	coordinates {(1, 79) (2, 29) (3, 20) (4, 35) (5, 65) (6, 52) (7, 21) (8, 44) (9, 64) (10, 0)};
    	\addlegendentry{ZS number}
    
        \end{axis}
    \end{tikzpicture}

\caption{Statistics of each task in WikiSQL and Spider.}
\label{fig:task}
\end{figure}
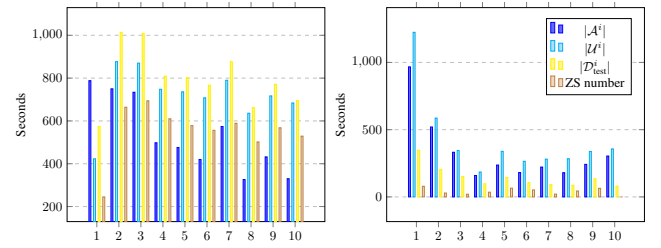

\section{Experiments}
\subsection{Experimental Setup}
\subsubsection{Datasets} To evaluate our proposed methods, we construct two task streams using the following two text-to-SQL datasets:
\textbf{WikiSQL}\footnote{\url{https://github.com/salesforce/WikiSQL}}~\cite{DBLP:journals/corr/abs-1709-00103} contains more than 20k tables collected from Wikipedia and 80,654 NLQ-SQL instances. Each instance corresponds to only a single table and the structure of the target SQL is relatively simple. 
\textbf{Spider}\footnote{\url{https://yale-lily.github.io//spider}}~\cite{DBLP:conf/emnlp/YuZYYWLMLYRZR18} contains 8,659 training instances across 146 RDs in total, and covers a wide range of domains, including flights, geography, movies, and more. Unlike WikiSQL, it instance corresponds to an RD containing multiple tables, and the target SQL may have a complex syntax.
\begin{table*} 
	\begin{center}
	{\caption{Experimental results for comparison with baselines.}\label{tab:overall_results}}
	\scalebox{0.95}{
		\begin{tabular}{lcccccccc}
			\toprule
			\multicolumn{1}{l}{\multirow{2}[1]{*}{\textbf{Method}}}
			&\multicolumn{4}{c}{\textbf{Spider}}&\multicolumn{4}{c}{\textbf{WikiSQL}}\\
		    \cmidrule(lr){2-5} \cmidrule(lr){6-9}
			
			&$\text{ACC}_{\text{a}}$ &$\text{ACC}_{\text{w}}$ 
			&$\text{BWT}$ 
			&$\text{FWT}$ 
			&$\text{ACC}_{\text{a}}$ &$\text{ACC}_{\text{w}}$
			&$\text{BWT}$ 
			&$\text{FWT}$ \\
			\cmidrule(lr){1-1} \cmidrule(lr){2-5} \cmidrule(lr){6-9}
			\textsc{Fine-tune} &47.5 &45.5 &-11.3 &38.4 &69.8 &69.2 &-1.5 &63.0 \\
			\cmidrule(lr){1-1} \cmidrule(lr){2-5} \cmidrule(lr){6-9}
			\textsc{Self-training}~\cite{DBLP:conf/acl/GoldwasserRCR11} &48.3 &46.6 &-10.9 &40.4 &70.4 &69.9 &-3.3 &63.5 \\
			\textsc{SETNet}~\cite{DBLP:conf/ecai/WangS0020} &47.8 &46.1 &-14.6 &41.6 &70.7 &70.2 &-2.1 &61.8 
			\\
			MST-SQL~\cite{DBLP:conf/ijcai/Guo0QWX22}  &49.6 &47.3 &-6.6 &40.7 &70.7 &70.1 &-1.7 &61.7 
			\\
			\cmidrule(lr){1-1} \cmidrule(lr){2-5} \cmidrule(lr){6-9}
			EWC~\cite{DBLP:journals/corr/KirkpatrickPRVD16} &48.3 &47.2 &-7.9 &38.4 &70.0 &69.6 &-2.0 &61.4  \\
			HAT~\cite{DBLP:conf/icml/SerraSMK18} &49.4 &47.7 &-8.4 &39.3 &70.0 &69.6 &\textbf{-1.4} &61.8  \\
			EMR~\cite{DBLP:conf/naacl/WangXYGCW19} &50.1 &49.1 &-3.2 &40.3 &71.1 &70.7 &-2.2 &63.1   \\
			EMAR~\cite{DBLP:conf/acl/HanDGLLLSZ20} &50.3 &49.6 &-4.6 &40.4 &70.8 &70.5 &-1.5 &62.7  \\
			APPER~\cite{DBLP:conf/emnlp/MiCZHF20} &50.7 &49 &-7.7 &40.0 &70.2 &69.9 &-3.0 &62.7  \\
			\textsc{Total-Recall}~\cite{DBLP:conf/emnlp/LiQH21} &53.4 &51.6 &-5.1 &40.3 &71.5 &71.1 &-2.1 &62.7  \\
			\cmidrule(lr){1-1} \cmidrule(lr){2-5} \cmidrule(lr){6-9}
		    \textsc{Vanilla} &53.9 &52.9 &-4.0 &40.6 &72.2 &71.9 &-2.0 &64.0  \\
		    \textsc{SFNet} &\textbf{56.0} &\textbf{53.6} &\textbf{-1.0} &\textbf{45.9} &\textbf{73.6} &\textbf{73.3} &-2.3 &\textbf{65.6}   \\
		    \cmidrule(lr){1-1} \cmidrule(lr){2-5} \cmidrule(lr){6-9}
		    \textsc{Oracle} (all tasks w/o Unsup.) &62.9 &63.4 &5.2 &48.7 &73.1 &72.7 &2.6 &64.2   \\
			\bottomrule
		\end{tabular}
		}
	\end{center}
\end{table*}
Following the problem formulation before, we divided each dataset into 10 tasks based on the domain of RD, and set the size of $\mathcal{A}^i$ for most tasks to less than 500. To provide a stable initialization for all compared baselines, we choose the task with the largest $\mathcal{A}^i$ as the first task. Refer to the practice in SSL~\cite{DBLP:conf/ijcai/Guo0QWX22}, we let most $\mathcal{U}^i$ be larger than $\mathcal{A}^i$. In addition, \cite{DBLP:conf/emnlp/YuZYYWLMLYRZR18} has pointed out that a good text-to-SQL method must have the ability to resolve zero-shot RDs. 
Thus, we guarantee that $\mathcal{D}^{i}_{\text{test}}$ has RDs or tables that are not seen in $\mathcal{A}^i$. Detailed statistics of our split datasets are illustrated in Figure \ref{fig:task}.

\subsubsection{Evaluation Metrics} 
Following previous works~\cite{DBLP:conf/naacl/WangXYGCW19,DBLP:conf/emnlp/LiQH21}, we adopt four metrics to evaluate the performance of the methods: 1) $\text{ACC}_{\text{a}} = \frac{1}{K} \sum_{i=1}^K acc_{i,K}$; 2) $\text{ACC}_{\text{w}} = acc_{\mathcal{D}_{\text{test}}^{(1:K)}}$; 3) $k < i$: $\text{BWT} = \frac{1}{K-1} \sum_{i=1}^{K-1} acc_{i,K} - acc_{i,i}$; 4) $\text{FWT} = \frac{1}{K-1} \sum_{i=2}^{K} acc_{i,i-1} - \bar{b}_{i}$, where $acc_{i,j}$ denotes the test accuracy on $\mathcal{D}^{i}_{\text{test}}$ after the training of $\mathcal{D}^{j}$ and $\bar{b}_i$ denotes the test accuracy on $\mathcal{D}^{i}_{\text{test}}$ at random initialization. The first two metrics mainly measure the comprehensive performance of parsers. The BWT measures forgetting of past tasks, while the FWT measures zero-shot performance on new tasks.

\subsubsection{Implementation Details} 
Our method ran on one Tesla A100 Super GPUs. We use pre-trained \textsc{Grappa}-Large as the encoder and \textit{K-medoids} as the clustering algorithm by default. The hyper-parameters were set as follow: (1) The maximum sizes of memory $\mathcal{M}_{\text{tea}}^i$ and $\mathcal{M}_{\text{stu}}^i$ were set to 30\% of $|\mathcal{D}^i_{\text{train}}|$. (2) The learning rate is set to $2 \times 10^{-5}$ for \textsc{Grappa} and $4 \times 10^{-4}$ for other modules. All our datasets and codes are publicly available\footnote{\url{https://github.com/Bahuia/SSCL-Text2SQL}}.

\subsubsection{Baselines} 
\textsc{Fine-tune} is an \textbf{naïve baseline} that uses only labeled data to fine-tune the model for the new task based on the previous model.
Self-training~\cite{DBLP:conf/acl/GoldwasserRCR11}, \textsc{SETNet}~\cite{DBLP:conf/ecai/WangS0020}, and \textsc{MST-SQL}~(Guo et al. 2022) composes the \textbf{SSL-only baselines} that also fine-tune the model based on the previous model but utilize both unlabeled and labeled data, where \textsc{SETNet} and \textsc{MST-SQL} apply mean-teacher and meta-learning to improve the generalization capability, respectively.
The \textbf{CL-only baselines} consists of EWC~\cite{DBLP:journals/corr/KirkpatrickPRVD16}, HAT~\cite{DBLP:conf/icml/SerraSMK18}, EMR~\cite{DBLP:conf/naacl/WangXYGCW19}, EMAR~\cite{DBLP:conf/acl/HanDGLLLSZ20}, APPER~\cite{DBLP:conf/emnlp/MiCZHF20}, and \textsc{Total-Recall}~\cite{DBLP:conf/emnlp/LiQH21}.
EWC uses regularization to constrain the learning of the current task. HAT applies a task-specific mask to guide each task. EMAR improves EMR with the constructed prototypes to avoid overfitting to memory. ARPER adds EWC regularization to the EMR loss and designs a priority-based sampling method. \textsc{Total-Recall} performs memory replay with a sampling method that balances the distribution of parsed actions and trains the semantic parser with a two-stage update strategy. Here we did not compare with GEM~\cite{DBLP:conf/nips/Lopez-PazR17} because its desired GPU memory is too large to run on our device. Finally, we set an approximate upper boundary \textsc{Oracle} that uses the \textit{full volume retraining}, i.e., for each task $\mathcal{D}^i$, the parser is trained by the combined data of all $\mathcal{A}^j$ ($1 \le j \le i$).

\subsection{Overall Results} 
The experimental results are shown in Table \ref{tab:overall_results}. The gap between \textsc{Fine-tune} and \textsc{Oracle} on WikiSQL are not as large as on Spider because the multiple tables and complex SQL syntax makes Spider more challenging. Despite the simplicity of the \textsc{Vanilla} process, it outperforms all the baselines in terms of $\text{ACC}_{\text{a}}$ and $\text{ACC}_{\text{w}}$.
More excitingly, our proposed \textsc{SFNet} further improves its $\text{ACC}_{\text{a}}$ by 1.4\% (WikiSQL) and 2.1\% (Spider), and achieves SOTA performance in almost all metrics on two datasets. Although \textsc{SFNet} does not perform best on WikiSQL in terms of BWT, it still shows the competitiveness with other methods and brings significant overall improvements on strong \textsc{Total Recall} by 2.1\% in $\text{ACC}_{\text{a}}$.
\begin{figure*}
\centering
    \begin{tikzpicture}[scale=0.47]
        \begin{axis}[
        ylabel={$\text{ACC}_{\text{a}}$}, 
        xtick=data,
        xticklabels={1, 2, 3, 4, 5, 6, 7, 8, 9, 10},
        tick align=inside, 
        legend pos=south west, 
        legend style={font=\small}, 
        ymajorgrids=true, 
        grid style=dashed]

            \addplot[smooth,mark=*,brown!30!white] plot coordinates { 
                (1, 64.82) (2, 56.62) (3, 52.48) (4, 50.56) (5, 52.07) (6, 53.08) (7, 52.47) (8, 47.76) (9, 46.86) (10, 47.52)
            };
        
            \addplot[smooth,mark=*,blue!30!white] plot coordinates {
                (1, 60.91) (2, 58.17) (3, 51.12) (4, 51.75) (5, 56.75) (6, 53.56) (7, 54.3) (8, 47.84) (9, 47.34) (10, 48.27)
            };
            
            \addplot[smooth,mark=*,cyan!30!white] plot coordinates {
            (1, 65.15) (2, 58.09) (3, 53.19) (4, 50.49) (5, 53) (6, 51.72) (7, 51.29) (8, 47.66) (9, 47.29) (10, 47.82)
            };
            
            \addplot[smooth,mark=*,pink] plot coordinates {
            (1, 62.54) (2, 56.79) (3, 47.66) (4, 47.77) (5, 54.41) (6, 51.44) (7, 51.5) (8, 46.5) (9, 47.09) (10, 49.62)
            };
            
            \addplot[smooth,mark=*,gray!30!white] plot coordinates {
                (1, 65.8) (2, 54.26) (3, 43.84) (4, 42.74) (5, 52.65) (6, 49.36) (7, 50.97) (8, 46.6) (9, 46.23) (10, 48.32)
            };
            
            
            \addplot[smooth,mark=pentagon,yellow] plot coordinates {
                (1, 60.26) (2, 59.31) (3, 53.57) (4, 48.3) (5, 54.4) (6, 52.12) (7, 54.38) (8, 45.38) (9, 44.47) (10, 49.44)
            };
            
           \addplot[smooth,mark=square,orange!30!white] plot coordinates {
            (1, 62.54) (2, 57.28) (3, 46.72) (4, 52.42) (5, 57.46) (6, 55.16) (7, 54.27) (8, 53.59) (9, 50.85) (10, 50.09)
            };
            
            \addplot[smooth,mark=square,orange] plot coordinates {
            (1, 62.54) (2, 61.43) (3, 50.86) (4, 54.53) (5, 53.74) (6, 52.56) (7, 53.28) (8, 49.16) (9, 47.95) (10, 50.34)
            };
            
            \addplot[smooth,mark=square,gray] plot coordinates {
               (1, 61.89) (2, 56.55) (3, 51.72) (4, 52.89) (5, 57.16) (6, 54.61) (7, 56.47) (8, 50.55) (9, 48.8) (10, 50.73)
            };
            
            \addplot[smooth,mark=square,brown] plot coordinates {
               (1, 62.1) (2, 60.12) (3, 50.19) (4, 52.62) (5, 57.67) (6, 55.09) (7, 53.16) (8, 53.54) (9, 53.45) (10, 53.4)
            };
            
            \addplot[smooth,mark=square,cyan] plot coordinates {
            (1, 57) (2, 54.75) (3, 54.54) (4, 54.9) (5, 58.52) (6, 57.72) (7, 57.03) (8, 55.08) (9, 55.98) (10, 53.91)
            };
            
            \addplot[smooth,mark=square,blue] plot coordinates {
            (1, 60.57) (2, 58.25) (3, 57.8) (4, 55.25) (5, 56.1) (6, 53.86) (7, 56.53) (8, 54.09) (9, 53.16) (10, 55.95) 
            };
            
        \end{axis}
    \end{tikzpicture}
    \begin{tikzpicture}[scale=0.47]
        \begin{axis}[
        ylabel={$\text{ACC}_{\text{w}}$}, 
        xtick=data,
        xticklabels={1, 2, 3, 4, 5, 6, 7, 8, 9, 10},
        tick align=inside, 
        legend pos=south west, 
        legend style={font=\small}, 
        ymajorgrids=true, 
        grid style=dashed]

            \addplot[smooth,mark=*,brown!30!white] plot coordinates { 
                (1, 64.82) (2, 54.49) (3, 49.45) (4, 46.22) (5, 49.86) (6, 50.95) (7, 49.29) (8, 45.63) (9, 44.4) (10, 45.45)
            };
        
            \addplot[smooth,mark=*,blue!30!white] plot coordinates {
                (1, 60.91) (2, 56.25) (3, 47.44) (4, 46.97) (5, 55.54) (6, 52.01) (7, 51.15) (8, 46.95) (9, 46.44) (10, 46.64)
            };
            
            \addplot[smooth,mark=*,cyan!30!white] plot coordinates {
            (1, 65.15) (2, 56.05) (3, 47.99) (4, 48.57) (5, 49.73) (6, 50.12) (7, 49.18) (8, 46.04) (9, 46.38) (10, 46.09)
            };
            
            \addplot[smooth,mark=*,pink] plot coordinates {
            (1, 60.91) (2, 56.25) (3, 47.44) (4, 46.97) (5, 55.54) (6, 52.01) (7, 51.15) (8, 46.95) (9, 46.44) (10, 46.64)
            };
            
            \addplot[smooth,mark=*,gray!30!white] plot coordinates {
                (1, 62.54) (2, 55.08) (3, 46.03) (4, 46.68) (5, 51.35) (6, 50.47) (7, 48.74) (8, 45.61) (9, 45.86) (10, 47.27)
            };
            
            
            \addplot[smooth,mark=pentagon,yellow] plot coordinates {
                (1, 60.26) (2, 58.01) (3, 49.27) (4, 46.72) (5, 53.78) (6, 49.29) (7, 50.93) (8, 44.41) (9, 42.11) (10, 47.73)
            };
            
            \addplot[smooth,mark=square,orange!30!white] plot coordinates {
            (1, 62.54) (2, 55.86) (3, 45.05) (4, 49.75) (5, 56.08) (6, 53.43) (7, 52.46) (8, 52.64) (9, 49.57) (10, 49.09)
            };
            
            \addplot[smooth,mark=square,orange] plot coordinates {
            (1, 62.54) (2, 60.74) (3, 45.42) (4, 50.25) (5, 52.97) (6, 51.77) (7, 50.93) (8, 48.07) (9, 48.13)
            (10, 49.64)};
            
            \addplot[smooth,mark=square,gray] plot coordinates {
               (1, 61.89) (2, 55.66) (3, 48.17) (4, 50.25) (5, 55.81) (6, 52.25) (7, 53.56) (8, 47.66) (9, 45.17) (10, 49)
            };
            
            \addplot[smooth,mark=square,brown] plot coordinates {
               (1, 62.1) (2, 58.2) (3, 48.9) (4, 52.1) (5, 56.49) (6, 54.73) (7, 51.15) (8, 51.51) (9, 51.47) (10, 51.55)
            };
            
            \addplot[smooth,mark=square,cyan] plot coordinates {
            (1, 57) (2, 53.13) (3, 50) (4, 52.44) (5, 56.08) (6, 56.5) (7, 55.42) (8, 53.35) (9, 54.16) (10, 52.91)
            };
            
            \addplot[smooth,mark=square,blue] plot coordinates {
            (1, 60.57) (2, 57.22) (3, 55.31) (4, 53.45) (5, 55.68) (6, 53.9) (7, 54.76) (8, 53.66) (9, 52.15) (10, 53.55)
            };
            
        \end{axis}
    \end{tikzpicture}
    \begin{tikzpicture}[scale=0.47]
        \begin{axis}[
        ylabel={BWT}, 
        xtick=data,
        tick align=inside, 
        legend style={font=\small, at={(1.48,0.95)}},
        ymajorgrids=true, 
        grid style=dashed]

            \addplot[smooth,mark=*,brown!30!white] plot coordinates { 
                (2, -18.89) (3, -16.76) (4, -13.93) (5, -10.92) (6, -6.54) (7, -7.77) (8, -11.39) (9, -11.76) (10, -11.31)
            };
        
            \addplot[smooth,mark=*,blue!30!white] plot coordinates {
                (2, -12.38) (3, -17.09) (4, -12.53) (5, -5.08) (6, -5.6) (7, -6.32) (8, -11.08) (9, -12.37) (10, -10.87)
            };
            
            \addplot[smooth,mark=*,cyan!30!white] plot coordinates {
            (2, -17.26) (3, -19.28) (4, -17.1) (5, -12.06) (6, -10.4) (7, -11.24) (8, -13.58) (9, -16.93) (10, -14.61)
            };
            
            \addplot[smooth,mark=*,pink] plot coordinates {
            (2, -14.33) (3, -24.81) (4, -17.48) (5, -6.36) (6, -6.83) (7, -7.25) (8, -10.6) (9, -10.12) (10, -6.6)
            };
            
            \addplot[smooth,mark=*,gray!30!white] plot coordinates {
                (2, -20.2) (3, -23.6) (4, -16.87) (5, -4.93) (6, -6.61) (7, -6.85) (8, -10.96) (9, -12.09) (10, -7.86)
            };
            
            
            \addplot[smooth,mark=pentagon,yellow] plot coordinates {
                (2, -7.49) (3, -13.59) (4, -14.55) (5, -6.39) (6, -6.59) (7, -6.36) (8, -14.01) (9, -14.56) (10, -8.4)
            };
            
            \addplot[smooth,mark=square,orange!30!white] plot coordinates {
               (2, -12.38) (3, -18.39) (4, -5.41) (5, -0.86) (6, -0.45) (7, -2.17) (8, -2.13) (9,-4.6) (10, -3.21)
            };
            
            \addplot[smooth,mark=square,orange] plot coordinates {
               (2, -4.6) (3, -19.77) (4, -11.06) (5, -9.79) (6, -6.43) (7, -6.7) (8, -8.48) (9,-9.18) (10, -4.63) 
            };
            
            \addplot[smooth,mark=square,gray] plot coordinates {
               (2, -9.78) (3, -13.27) (4, -8.41) (5, -3.78) (6, -4.09) (7, -4.07) (8, -8.83) (9, -10.35) (10, -7.69)
            };
            
            \addplot[smooth,mark=square,brown] plot coordinates {
            (2, -8.61) (3, -6.15) (4, -4.46) (5, -1.88) (6, -2.69) (7, -4.51) (8, -3.95) (9, -5.21) (10, -5.05)
            };
            
            \addplot[smooth,mark=square,cyan] plot coordinates {
            (2, -10.42) (3, -9.04) (4, -3.69) (5, -1.07) (6, -0.08) (7, -1.32) (8, -2.26) (9, -1.65) (10, -4.02)
            };
            
            \addplot[smooth,mark=square,blue] plot coordinates {
            (2, -7.49) (3, -6.19) (4, -5.26) (5, -3.72) (6, -3.5) (7, -1.77) (8, -2.07) (9, -3.47) (10, -0.95)
            };
            
        \end{axis}
    \end{tikzpicture}
    \begin{tikzpicture}[scale=0.47]
        \begin{axis}[
        ylabel={FWT}, 
        xtick=data,
        xticklabels={1, 2, 3, 4, 5, 6, 7, 8, 9, 10},
        tick align=inside, 
        legend style={font=\small, at={(1.48,0.95)}}, 
        ymajorgrids=true, 
        grid style=dashed]

            \addplot[smooth,mark=*,brown!30!white] plot coordinates { 
                (2, 31.22) (3, 39.14) (4, 40.38) (5, 40.97) (6, 41.08) (7, 41.7) (8, 40.37) (9, 40.24) (10, 38.39)
            };
            \addlegendentry{\textsc{Fine-tune}}
        
            \addplot[smooth,mark=*,blue!30!white] plot coordinates {
                (2, 30.24) (3, 40.12) (4, 40.35) (5, 39.92) (6, 40.05) (7, 41.58) (8, 40.87) (9, 42.53) (10, 40.43)
            };
            \addlegendentry{\textsc{Self-training}}
            
            \addplot[smooth,mark=*,cyan!30!white] plot coordinates {
            (2, 31.71) (3, 43.79) (4, 44.16) (5, 43.64) (6, 43.59) (7, 44.53) (8, 42.6) (9, 43.63) (10, 41.61)
            };
            \addlegendentry{\textsc{SETNet}}
            
            \addplot[smooth,mark=*,pink] plot coordinates {
            (2, 29.27) (3, 38.16) (4, 38.37) (5, 39.29) (6, 39.74) (7, 40.82) (8, 40.42) (9, 42.13) (10, 40.08)
            };
            \addlegendentry{\textsc{MST-SQL}}
            
            \addplot[smooth,mark=*,gray!30!white] plot coordinates {
                (2, 29.76) (3, 39.88) (4, 36.79) (5, 37.59) (6, 37.43) (7, 39.15) (8, 37.99) (9, 36.12) (10, 38.36)
            };
            \addlegendentry{EWC}
            
            
            \addplot[smooth,mark=pentagon,yellow] plot coordinates {
            (2, 27.32) (3, 35.72) (4, 38.78) (5, 40.46) (6, 40.67) (7, 40.61) (8, 40.44) (9, 41.53) (10, 39.34)
            };
            \addlegendentry{HAT}
            
            \addplot[smooth,mark=square,orange!30!white] plot coordinates {
             (2, 30.24) (3, 38.65) (4, 39.37) (5, 39.53) (6, 39.36) (7, 40.76) (8, 39.98) (9, 41.94) (10, 40.31)
            };
            \addlegendentry{EMR}
            
            \addplot[smooth,mark=square,orange] plot coordinates {
            (2, 31.21) (3, 37.67) (4, 38.72) (5, 42.31) (6, 40.64) (7, 42.08) (8, 41.5) (9, 42.26) (10, 40.39)
            };
            \addlegendentry{EMAR}
            
            \addplot[smooth,mark=square,gray] plot coordinates {
               (2, 29.76) (3, 36.94) (4, 37.55) (5, 39.02) (6, 39.33) (7, 41.48) (8, 40.59) (9, 41.87) (10, 40.04)
            };
            \addlegendentry{APPER}
            
            \addplot[smooth,mark=square,brown] plot coordinates {
               (2, 29.76) (3, 35.47) (4, 35.89) (5, 37.61) (6, 38.01) (7, 39.88) (8, 39.02) (9, 39.67) (10, 40.29)
            };
            \addlegendentry{\textsc{Total-Recall}}
            
            \addplot[smooth,mark=square,cyan] plot coordinates {
            (2, 27.32) (3, 37.19) (4, 41.12) (5, 41.36) (6, 41.2) (7, 43.78) (8, 42.96) (9, 42.3) (10, 40.63)
            };
            \addlegendentry{\textsc{Vanilla}}
            
            \addplot[smooth,mark=square,blue] plot coordinates {
            (2, 31.71) (3, 45.27) (4, 47.86) (5, 47.97) (6, 46.3) (7, 47.29) (8, 45.76) (9, 46.81) (10, 45.85)
            };
            \addlegendentry{\textsc{SFNet}}
            
        \end{axis}
    \end{tikzpicture}
\caption{$\text{ACC}_{\text{a}}$, $\text{ACC}_{\text{w}}$, BWT, FWT till the seen tasks on Spider after learning on each task sequentially.}
\label{fig:acc_till_task}
\end{figure*}
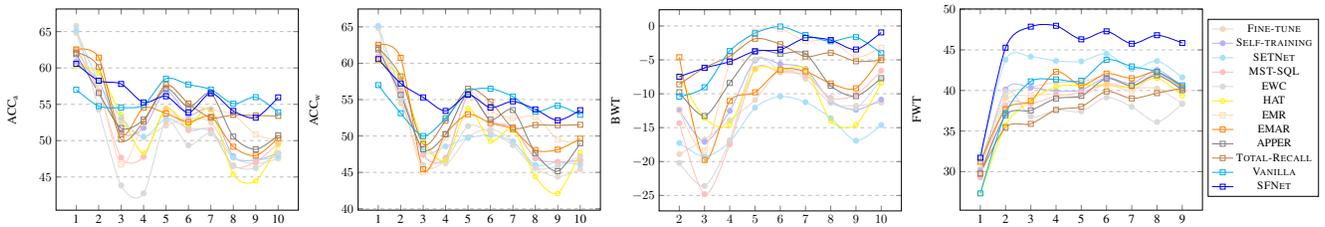
The SSL-only methods improve the overall performance of \textsc{Fine-tune} using the information provided by unsupervised data, while they perform poorly in fighting against the CF (-10.9\% and -14.6\%). MST-SQL achieves better results, probably because of its meta-learning that captures common features across tasks.
The replay-based EMR, EMAR, and \textsc{Total Recall} perform better than other CL-only methods on BWT, which proves that past instances may be more important to overcome the CF of the text-to-SQL parser. Although \textsc{Total Recall} benefits from its fine-grained semantic parsing sampling algorithm to achieve a best results of the baselines, its results are still limited by ignoring unsupervised information. Unlike them, our \textsc{SFNet} deeply integrates SSL and CL and leads them to reinforce each other, thus achieving overall excellent results.

\begin{table} 
	\begin{center}
	{\caption{Ablation study results of \textsc{SFNet}}\label{tab:ablation_test}}
	\scalebox{0.73}{
		\begin{tabular}{lcccccccc}
			\toprule
			\multicolumn{1}{l}{\multirow{2}[1]{*}{\textbf{Setting}}}
			&\multicolumn{4}{c}{\textbf{Spider}}&\multicolumn{4}{c}{\textbf{WikiSQL}}\\
		    \cmidrule(lr){2-5} \cmidrule(lr){6-9}
			&A &W &B &F 
			&A &W &B &F  \\
            \cmidrule(lr){1-1}
			\cmidrule(lr){2-5} \cmidrule(lr){6-9}
			\textsc{SFNet} &\textbf{56.0} &\textbf{54.1} &-1.0 &\textbf{45.9} &\textbf{73.6} &\textbf{73.3} &-2.3 &\textbf{65.6} \\
			\cmidrule(lr){1-1}
			\cmidrule(lr){2-5} \cmidrule(lr){6-9}
		    w/o $\mathcal{F}^i_{\text{tea}}$ &54.5 &52.5 &-4.0 &42.9 &72.0 &71.6 &\textbf{-1.3} &63.7  \\
		    w/o $\mathcal{F}^i_{\text{stu}}$ &50.4 &48.2 &-7.2 &45.1 &71.6 &72.7 &-2.0 &63.5 \\
		    Combine $\mathcal{F}^i_{\text{tea}}$ \& $\mathcal{F}^i_{\text{stu}}$ &54.8 &53.0 &-3.0 &43.6 &71.7 &71.3 &-2.0 &62.9 \\
		    w/o PS &53.7 &52.9 &-1.1 &41.8 &70.4 &70.0 &-4.6 &62.6  \\
		    RS only using $\mathcal{A}^i$ &54.2 &51.4 &\textbf{-0.9} &43.1 &71.4 &71.0 &-2.8 &62.7  \\
			\bottomrule
		\end{tabular}
		}
	\end{center}
\end{table}


\subsection{Detailed Results and Analysis}

\subsubsection{Results till the Seen Tasks} 
Figure \ref{fig:acc_till_task} shows the results of four metrics till the seen tasks on Spider after learning one each task. We can see that our proposed \textsc{SFNet} (blue) is always more stable than the other baselines in all metrics and this stability becomes more pronounced as the number of tasks grows. In terms of BWT and FWT, the improvement brought by \textsc{SFNet} is more significant, which proves the effectiveness of our fusion manner to SSL and CL, not only alleviating forgetfulness of past tasks, but also improving the generalization capability for zero-shot RDs. Notice that the performance of almost all methods in BWT improves slightly as the number of tasks increases. This may be related to the combined generalization of text-to-SQL tasks, i.e., the parser might resolve the few instances that were previously incorrectly predicted with the SQL fragments learned in the new task. Building on this foundation, our \textsc{SFNet} further exacerbates this trend using PS.

\subsubsection{Ablation Test}

\begin{table} 
	\begin{center}
	{\caption{$\text{ACC}_{\text{a}}$ (A), $\text{ACC}_{\text{w}}$ (W), BWT (B), and FWT (F) of \textsc{SFNet} with different sampling strategies.}\label{tab:sampling}}
	\scalebox{0.75}{
		\begin{tabular}{lccccccccc}
			\toprule
			\multicolumn{1}{l}{\multirow{2}[1]{*}{\textbf{Methods}}}
			&\multicolumn{4}{c}{\textbf{Spider}}&\multicolumn{4}{c}{\textbf{WikiSQL}}\\
		    \cmidrule(lr){2-5} \cmidrule(lr){6-9}
			
			&A &W &B &F 
			&A &W &B &F\\

			\cmidrule(lr){1-1} \cmidrule(lr){2-5} \cmidrule(lr){6-9}
			\textsc{Random}  &53.2 &52.7 &-2.2 &41.3 &72.2 &71.8 &-3.0 &62.9 \\
			\textsc{Schema} Sim.  &54.3 &53.2 &\textbf{-1.0}  &45.0 &72.8 &72.4 &-2.0 &63.9 \\
			\cmidrule(lr){1-1} \cmidrule(lr){2-5} \cmidrule(lr){6-9}
		    \textsc{Random}  &52.3 &51.4 &-3.3 &41.5 &71.5 &71.1 &-1.8 &64.6 \\
		    \textsc{FSS}  &52.6 &51.9 &-2.0 &42.0 &72.3 &72.1 &-2.0 &64.7 \\
		    \textsc{Prior}  &52.6 &51.1 &-4.0 &42.0 &71.9 &71.3 &\textbf{-1.5} &64.1 \\
		    \textsc{Balance}  &52.0 &49.6 &-2.4 &41.1 &71.9 &71.4 &-2.8 &64.3 \\
		    \textsc{LFS}  &52.4 &51.2 &-6.3 &40.4 &72.1 &71.6 &-2.2 &63.5 \\
		    \textsc{DLFS}  &52.8 &51.9 &-1.5 &40.5 &72.2 &71.9 &-2.7 &64.7 \\
		    \textsc{Schema} Clus.  &54.9 &53.3 &-1.1 &43.8 &72.3 &71.8 &-2.1 &64.1 \\
		    \cmidrule(lr){1-1} \cmidrule(lr){2-5} \cmidrule(lr){6-9}
		    Dual Sampling  &\textbf{55.7} &\textbf{54.1} &\textbf{-1.0} &\textbf{45.9} &\textbf{73.6} &\textbf{73.3} &-2.3 &\textbf{65.6} \\
			\bottomrule
		\end{tabular}
		}
	\end{center}
\end{table}

To explore the contributions of each component of our proposed \textsc{SFNet}, we compared the performance of the following settings:
\begin{itemize}
    \item \textbf{w/o $\mathcal{F}^i_{\text{tea}}$} We remove $\mathcal{F}^i_{\text{tea}}$ and in-task ST to verify the contribution of SSL;
    \item \textbf{w/o $\mathcal{F}^i_{\text{stu}}$} We remove $\mathcal{F}_{\text{stu}}$ and cross-task EMR, and only use $\mathcal{F}_{\text{tea}}$ to predict, to evaluate the contribution of CL;
    \item \textbf{Combine $\mathcal{F}^i_{\text{tea}}$ \& $\mathcal{F}^i_{\text{stu}}$} We merged $\mathcal{F}_{\text{tea}}$ and $\mathcal{F}_{\text{stu}}$ into a single parser to assess the necessity of the TS framework.
    \item \textbf{w/o PS}  We remove the prompt sampling to evaluate the improvements brought to SSL by past information.
    \item \textbf{RS only using $\mathcal{A}^i$}; We review the sampling using only the labeled set $\mathcal{A}^i$ to assess the improvement of CL by pseudo-supervision information.
\end{itemize}

Table \ref{tab:ablation_test} shows the $\text{ACC}_{\text{a}}$ (A), $\text{ACC}_{\text{w}}$ (W), BWT (B), and FWT (F) of  different settings. Our \textsc{SFNet} equipped with all modules performs best in terms of $\text{ACC}_{\text{a}}$ and FWT. Its non-negligible improvements in overall performance allow us to overlook the minor shortcomings in BWT.
By removing $\mathcal{F}^i_{\text{tea}}$, $\text{ACC}_{\text{a}}$ decreases by 1.6\% on WikiSQL and 1.5\% for Spider, which proves that the value of the unsupervised information. Discarding $\mathcal{F}^i_{\text{stu}}$ leads to a decrease in terms of $\text{ACC}_{\text{a}}$ (-2.0\% \& -5.4\%) because of the forgetting of previous tasks. The performance drop (-1.9\% \& -1.2\%) brought by abandoning the TS framework demonstrates the necessity of handling SSL and CL separately. On both datasets, removing PS results in an absolute decrease in all metrics. This reveals that the relevant information from past tasks can prompt the SSL process of current task. 
In particular, the drop in FWT on Spider (-4.1\%) is more significant, probably due to the fact that Spider can provide information not only on the RD schema but also on the SQL structure.
The degradation in $\text{ACC}_{\text{a}}$ aspect when using only $\mathcal{A}^i$ and not $\mathcal{P}^i$ in the RS process reflects that the SSL results can also contribute to the CL process.

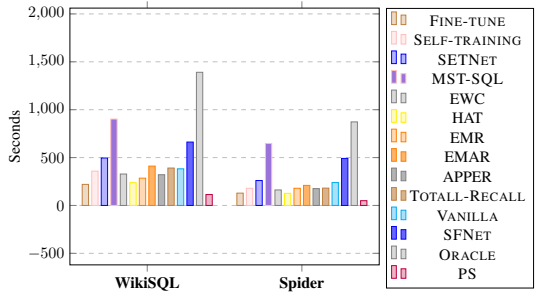
\begin{figure}
\centering
    	
    	
    	
    
    	
    	
    	
    
    \begin{tikzpicture}[scale=0.6]
        \begin{axis}[
        ybar,
        bar width=4pt,
        enlargelimits=0.5,
        legend style={at={(1.5,1)}},
        ylabel=Seconds,
        symbolic x coords={\textbf{WikiSQL}, \textbf{Spider}},
    	xtick=data,
        ymajorgrids=true, 
        grid style=dashed]
    	
        \addplot[draw=brown, fill=brown!30!white]
    	coordinates {(\textbf{WikiSQL}, 219.93454) (\textbf{Spider}, 128.77174)};
    	\addlegendentry{\textsc{Fine-tune}}
    	
    	\addplot[draw=pink, fill=pink!30!white] 
    	coordinates {(\textbf{WikiSQL}, 358.34478) (\textbf{Spider},179.23629) };
    	\addlegendentry{\textsc{Self-training}}
    
        \addplot[draw=blue, fill=blue!30!white] 
    	coordinates {(\textbf{WikiSQL}, 495.50402) (\textbf{Spider}, 260.36957) };
    	\addlegendentry{\textsc{SETNet}}
    
        \addplot[draw=pink, fill=pink!60!blue]  
    	coordinates {(\textbf{WikiSQL}, 902.34586) (\textbf{Spider},647.24004)};
    	\addlegendentry{\textsc{MST-SQL}}
    	
    	\addplot[draw=gray, fill=gray!30!white] 
    	coordinates {(\textbf{WikiSQL}, 328.20781) (\textbf{Spider},162.00177) };
    	\addlegendentry{EWC}
    	
    	
    	\addplot[draw=yellow, fill=yellow!30!white]  
    	coordinates {(\textbf{WikiSQL}, 237.87114) (\textbf{Spider}, 126.44652) };
    	\addlegendentry{HAT}
    	
    	\addplot[draw=orange, fill=orange!30!white]   
    	coordinates {(\textbf{WikiSQL}, 284.96174) (\textbf{Spider},179.60856) };
    	\addlegendentry{EMR}
    	
    	\addplot[draw=orange, fill=orange!60!white]   
    	coordinates {(\textbf{WikiSQL}, 411.00086) (\textbf{Spider},209.56378) };
    	\addlegendentry{EMAR}
    	
    	\addplot[draw=gray, fill=gray!60!white]   
    	coordinates {(\textbf{WikiSQL}, 320.9023) (\textbf{Spider},175.97007) };
    	\addlegendentry{APPER}
    	
    	\addplot[draw=brown, fill=brown!60!white]   
    	coordinates {(\textbf{WikiSQL}, 390.72956) (\textbf{Spider},181.15107) };
    	\addlegendentry{\textsc{Totall-Recall}}
    	
    	\addplot[draw=cyan, fill=cyan!30!white]   
    	coordinates {(\textbf{WikiSQL},382.62985) (\textbf{Spider},239.16986) };
    	\addlegendentry{\textsc{Vanilla}}
    	
    	\addplot[draw=blue, fill=blue!60!white]   
    	coordinates {(\textbf{WikiSQL},662.00749) (\textbf{Spider},491.09192) };
    	\addlegendentry{\textsc{SFNet}}
    	
    	\addplot[draw=gray, fill=gray!30!white]   
    	coordinates {(\textbf{WikiSQL},1390.44522) (\textbf{Spider},872.55652) };
    	\addlegendentry{\textsc{Oracle}}
    	
    	\addplot[draw=purple, fill=purple!30!white]   
    	coordinates {(\textbf{WikiSQL},114.01618) (\textbf{Spider},49.88979) };
    	\addlegendentry{\textsc{PS}}
    
    \end{axis}
    \end{tikzpicture}

\caption{Average training time of each method on one task.}
\label{fig:time}
\end{figure}

\subsubsection{Impact of Sampling Methods} 
To further evaluate our proposed PS and RS, we replace them with some other existing sampling strategies. For prompt sampling, we replace it \textsc{Random} and \textsc{Schema} Sim., where the former randomly samples training instances and the latter only selects the top-$k$ instances by schema similarity $\omega(x)$ without using $d_{\text{stru}}$ to clustering. For review sampling, we replace it with the sampling strategies widely-used in CL, including FSS~\cite{DBLP:conf/naacl/WangXYGCW19,DBLP:conf/nips/AljundiLGB19}, LFS~\cite{DBLP:conf/emnlp/LiQH21}, \textsc{Prior}~\cite{DBLP:conf/emnlp/MiCZHF20}, and \textsc{Balance}~\cite{DBLP:conf/pakdd/JianYZ22}, and  our defined \textsc{Schema} Clus., which performs clustering only uses $d_{\text{sch}}$. Here, \textsc{Schema} Sim. and \textsc{Schema} Clus. are used to evaluate the contribution of the diverse SQL structures to the performance.
From Table \ref{tab:sampling} we can see that our dual sampling outperforms all other baselines on both datasets in terms of $\text{ACC}_{\text{a}}$ and $\text{ACC}_{\text{w}}$. More importantly, the improvements over \textsc{Schema} Sim. and \textsc{Schema} Clus. in terms of FWT reveals the fact that for text-to-SQL, diverse SQL structures in the training data are useful for the parser to generate unseen SQL programs.


\subsubsection{Training Time Analysis} 
The average training times of different methods on each task are depicted in Figure \ref{fig:time}. EMR and EMAR are slower than other CL-only methods like HAT and EWC because they need to replay the past instances during the training. Similarly, SSL-only methods predicts pseudo-labels on unsupervised data to augment the training data, and thus also require longer training time. In addition, the training process of MST-SQL is exceptionally time-consuming because it requires the construction of multiple sets of meta-learning tasks with possible data duplication. Notably, our \textsc{SFNet} contains the training process for both models and utilizes both unsupervised and replayed instances, while it takes only half of the time used by \textsc{Oracle} that does not uses unsupervised data. Moreover, although our PS accesses the full amount of past data each time, the fast hashing strategy makes it take only about 1/6 of the total time \textsc{SFNet} uses.


\section{Related Work}
\noindent \textbf{Text-to-SQL}
Research on text-to-SQL can be roughly divided into three directions. The first one is the single-table task represented by \textsc{WikiSQL}~\cite{DBLP:journals/corr/abs-1709-00103}, whose target SQL programs contain only simple syntaxes. The SOTA methods~\cite{DBLP:journals/corr/abs-1902-01069,DBLP:conf/aaai/ChenG0QQWL21,DBLP:conf/naacl/XuWWWWD22} of \textsc{WikiSQL} typically treat the problem as several subtasks and leverage \textit{multi-task learning} to solve them. The second direction is a cross-domain multi-table scenario represented by \textsc{Spider}~\cite{DBLP:conf/emnlp/YuZYYWLMLYRZR18}. Its target programs cover a variety of complicated SQL syntaxes including \texttt{GROUP BY} and nested queries, to better meet the needs of real-world applications. In this scenario, most SOTA parsers~\cite{DBLP:conf/acl/GuoZGXLLZ19,DBLP:conf/acl/WangSLPR20,DBLP:conf/acl/CaoC0ZZ020} apply top-down grammar-based decoding, which is consistent with our work. The last one is a conversational task, such as \textsc{SParC}~\cite{DBLP:conf/acl/YuZYTLLELPCJDPS19} and \textsc{CoSQL}~\cite{DBLP:conf/emnlp/YuZELXPLTSLJYSC19}, forcing the parser to learn to consider the context information when generating SQL in a multi-turn dialogue~\cite{DBLP:conf/emnlp/ZhangYESXLSXSR19,DBLP:conf/acl/ZhengWDWL22}.

\noindent \textbf{SSL in Semantic Parsing} 
Multiple classical methods have been applied to address the challenge of lack of annotation for semantic parsing, such as SVM~\cite{DBLP:conf/naacl/KateM07}, \textit{self-training}~\cite{DBLP:conf/acl/GoldwasserRCR11}, \textit{dual learning}~\cite{DBLP:conf/acl/ChenCLCMWY21}, \textit{auto-encoder}~\cite{DBLP:conf/acl/NeubigZYH18}, and \textit{mean-teacher}~\cite{DBLP:conf/ecai/WangS0020}. We finally adopted self-training in the proposed solutions for simplicity. Unlike previous methods, we propose relevance sampling to utilize the continual scenario to boost the performance of SSL.

\noindent \textbf{CL in Semantic Parsing} 
There is relatively little work that applies CL to semantic parsing. \cite{lialin2020update} and \cite{DBLP:conf/emnlp/LiQH21} apply EWC~\cite{DBLP:journals/corr/KirkpatrickPRVD16} and EMR~\cite{DBLP:conf/naacl/WangXYGCW19}, respectively, to handle the task stream of traditional semantic parsing benchmarks. Different from them, our method focuses on text-to-SQL with larger application scenarios and further argument the memories in the continual process using unsupervised data.

\noindent \textbf{Semi-supervised Continual Learning} 
Recent SSCL methods~\cite{DBLP:journals/corr/abs-2110-01856,DBLP:journals/corr/abs-2201-09196,DBLP:conf/ijcnn/SmithBHK21,DBLP:conf/cvpr/WangYLHL021} have focused on image classification tasks. They are either based on backbone models widely used in Computer Vision (e.g., \textit{generative adversarial networks}) or on gradient prediction which is computationally expensive and therefore not suitable for Natural Language Processing tasks with large-scale pre-trained models. In contrast, our research is dedicated to obtaining a simple and efficient method for text-to-SQL within the allowed time-space overhead.

\section{Conclusion}
In this paper, we presented two methods that integrates \textit{semi-supervised learning} (SSL) and \textit{continual learning} (CL) to address the problem of supervision-limited text-to-SQL task stream. 
The first approach combines self-training and episodic memory replay to enhance supervision while balancing training efficiency and performance of the overall task stream.
The improved method \textsc{SFNet} drives the intrinsic connection between CL and SSL by using in-memory past information to help current SSL, while adding high-quality pseudo instances in memory to improve future replay. 
The experiments on two benchmarks shows that indicate that our method provide a promising way for  supervision-limited text-to-SQL task stream. In future work, we will try to introduce \textit{prompt learning} into the continual process to augment supervision and alleviate forgetting.

\bibliography{aaai22}
\end{document}